\documentclass[conference]{IEEEtran}
\IEEEoverridecommandlockouts
\usepackage{cite}
\usepackage{amsmath,amssymb,amsfonts}
\usepackage{algorithmic}
\usepackage{graphicx}
\usepackage{textcomp}
\usepackage{xcolor}
\def\BibTeX{{\rm B\kern-.05em{\sc i\kern-.025em b}\kern-.08em
    T\kern-.1667em\lower.7ex\hbox{E}\kern-.125emX}}
\begin{document}

\title{Enhancing AI-Driven Education: Integrating Cognitive Frameworks, Linguistic Feedback Analysis, and Ethical Considerations for Improved Content Generation}

%\author{\IEEEauthorblockN{Anonymous Authors}}
%%
%%%
\author{\IEEEauthorblockN{Antoun Yaacoub}
\IEEEauthorblockA{\textit{Learning,  Data  and  Robotics  (LDR)  } \\
\textit{ESIEA Lab, ESIEA}\\
Paris, France \\
antoun.yaacoub@esiea.fr}
\and
\IEEEauthorblockN{Sansiri Tarnpradab}
\IEEEauthorblockA{\textit{Department of Computer Engineering} \\
\textit{King Mongkut's University}\\ \textit{of Technology Thonburi}\\
Bangkok, Thailand \\
sansiri.tarn@mail.kmutt.ac.th}
\and
\IEEEauthorblockN{Phattara Khumprom}
\IEEEauthorblockA{\textit{Graduate School of Management}\\ \textit{and Innovation} \\
\textit{King Mongkut's University}\\ \textit{of Technology Thonburi}\\
Bangkok, Thailand \\
phattara.khum@kmutt.ac.th}
\and
\IEEEauthorblockN{Zainab Assaghir}
\IEEEauthorblockA{\textit{\hspace{2cm}Faculty  of  Science\hspace{2cm}} \\
\textit{Lebanese  University}\\
Beirut,  Lebanon \\
zainab.assaghir@ul.edu.lb}
\and
\IEEEauthorblockN{Lionel Prevost}
\IEEEauthorblockA{\textit{Learning,  Data  and  Robotics  (LDR)  } \\
\textit{ESIEA Lab, ESIEA}\\
Paris, France \\
lionel.prevost@esiea.fr}
\and
\IEEEauthorblockN{J\'er\^ome  Da-Rugna}
\IEEEauthorblockA{\textit{Learning,  Data  and  Robotics  (LDR)} \\
\textit{ESIEA Lab, ESIEA}\\
Paris, France \\
jerome.darugna@esiea.fr}
}

\maketitle

\begin{abstract}
Artificial intelligence (AI) is rapidly transforming education, presenting unprecedented opportunities for personalized learning and streamlined content creation. However, realizing the full potential of AI in educational settings necessitates careful consideration of the quality, cognitive depth, and ethical implications of AI-generated materials. This paper synthesizes insights from four related studies to propose a comprehensive framework for enhancing AI-driven educational tools. We integrate cognitive assessment frameworks (Bloom's Taxonomy and SOLO Taxonomy), linguistic analysis of AI-generated feedback, and ethical design principles to guide the development of effective and responsible AI tools. We outline a structured three-phase approach encompassing cognitive alignment, linguistic feedback integration, and ethical safeguards. The practical application of this framework is demonstrated through its integration into OneClickQuiz, an AI-powered Moodle plugin for quiz generation. This work contributes a comprehensive and actionable guide for educators, researchers, and developers aiming to harness AI's potential while upholding pedagogical and ethical standards in educational content generation.
\end{abstract}


\section{Introduction}
Artificial Intelligence (AI) is revolutionizing various aspects of modern life, and education is no exception [1]. From personalized learning platforms to automated assessment systems, AI holds the potential to transform teaching methodologies, content creation, and student learning experiences [1, 2]. In particular, recent advancements in generative AI have opened exciting new avenues for creating diverse and engaging educational materials, such as quizzes, assignments, and interactive learning modules [3, 4].

However, realizing the full potential of AI in education is not without its challenges. Simply automating content generation is insufficient; it is crucial to ensure that AI-driven educational tools are pedagogically sound, cognitively stimulating, and ethically responsible [5, 6]. One critical aspect is aligning AI-generated content with established cognitive assessment frameworks, such as the Structure of the Observed Learning Outcome (SOLO) Taxonomy and Bloom's Taxonomy, which provide structured approaches to categorizing learning outcomes and cognitive skills [7, 8]. Furthermore, it is essential to carefully analyze the feedback provided by AI systems to ensure it is linguistically appropriate, supportive, and conducive to learning [9, 10]. Finally, ethical considerations surrounding bias, fairness, and transparency must be addressed to ensure that AI tools promote equitable and inclusive education for all students [11, 12].

To address these multifaceted challenges, this paper synthesizes insights from four related studies conducted by our research group:
\begin{itemize}
\item  Study 1 (OneClickQuiz): Presented OneClickQuiz, an AI-powered Moodle plugin for automated quiz generation, showcasing its potential to streamline educational workflows and enhance student engagement [13].
\item  Study 2 (Bloom's Taxonomy): Evaluated the integration of Bloom's Taxonomy into AI-driven question generation, highlighting the potential of advanced models in classifying questions across Bloom's levels [14].
\item Study 3 (SOLO Taxonomy): Explored the alignment of AI-generated tasks with the SOLO Taxonomy, demonstrating the effectiveness of advanced models like DistilBERT in classifying questions across different cognitive levels [15].
\item  Study 4 (Feedback Analysis): Analyzed the linguistic characteristics of AI-generated feedback in multiple-choice questions, revealing significant interactions between feedback tone and question difficulty [16].
\end{itemize}

Building upon these individual contributions, this paper proposes a comprehensive framework for enhancing AI-driven education. We integrate cognitive assessment frameworks (SOLO and Bloom's), linguistic analysis of AI-generated feedback, and ethical design principles to guide the development of more effective and responsible AI tools. Specifically, we address the following research question:

\textit{How can we synthesize cognitive assessment frameworks (Bloom's and SOLO), insights from linguistic analysis of AI-generated feedback, and ethical design principles to develop more effective and responsible AI-driven educational tools for content creation and assessment?}

To answer this question, we outline a three-phase approach involving cognitive alignment, linguistic feedback integration, and ethical safeguards. We demonstrate the practical application of this framework by showcasing enhancements to OneClickQuiz, an AI-powered Moodle plugin for quiz generation. This work contributes a comprehensive and actionable framework for educators, researchers, and developers seeking to harness the power of AI while upholding pedagogical and ethical standards in educational content generation.

The remainder of this paper is structured as follows: Section 2 provides a more in-depth review of background concepts, Section 3 presents our three-phase framework, Section 4 demonstrates the utility of the framework through OneClickQuiz, Section 5 presents our findings, and Section 6 provides our conclusions and implications for further study.

\section{Background}
To effectively leverage AI for educational content generation, it is essential to understand the theoretical foundations that underpin learning and assessment. This section provides a detailed overview of relevant cognitive assessment frameworks, discusses the importance of linguistic analysis of AI-generated feedback, and outlines the key ethical considerations that must guide the design and implementation of AI-driven educational tools.
\subsection{Cognitive Assessment Frameworks: Bloom's and SOLO Taxonomies}
Cognitive assessment frameworks provide structured approaches to categorizing learning outcomes and evaluating the complexity of student understanding. These frameworks serve as valuable tools for educators and designers seeking to create assessments that align with specific learning objectives and promote higher-order thinking. In this paper, we focus on two prominent frameworks: Bloom's Taxonomy and the Structure of the Observed Learning Outcome (SOLO) Taxonomy.
\subsubsection{Bloom's Taxonomy}
Bloom's Taxonomy, originally developed by Bloom \textit{et al.} (1956) and later revised by Anderson and Krathwohl (2001), offers a hierarchical classification of cognitive learning objectives [8]. The revised taxonomy outlines six cognitive process categories:
\begin{itemize}
\item \textbf{Remember}: Retrieving relevant knowledge from long-term memory.
\item \textbf{Understand}: Constructing meaning from instructional messages, including oral, written, and graphic communication.
\item \textbf{Apply}: Carrying out or using a procedure in a given situation.
\item \textbf{Analyze}: Breaking material into constituent parts and determining how the parts relate to one another and to an overall structure or purpose.
\item \textbf{Evaluate}: Making judgments based on criteria and standards.
\item \textbf{Create}: Putting elements together to form a coherent or functional whole; reorganizing elements into a new pattern or structure.
\end{itemize}

Bloom's Taxonomy is widely used to guide curriculum development, instructional design, and assessment. It provides a framework for creating learning objectives that promote higher-order thinking and ensures that assessments align with learning goals.

\subsubsection{Structure of the Observed Learning Outcome (SOLO) Taxonomy}
The SOLO Taxonomy, developed by Biggs and Collis (1982), provides a hierarchical model for classifying learning outcomes based on their structural complexity [7]. The taxonomy outlines five distinct levels of understanding:
\begin{itemize}
\item \textbf{Prestructural}: The learner has no understanding of the task or subject matter.
\item \textbf{Unistructural}: The learner can identify one relevant aspect of the task.
\item \textbf{Multistructural}: The learner can identify several relevant aspects but cannot integrate them.
\item \textbf{Relational}: The learner can integrate different aspects into a coherent whole, demonstrating an understanding of relationships.
\item \textbf{Extended Abstract}: The learner can generalize the integrated whole to new situations and make predictions based on abstract thinking.
\end{itemize}

The SOLO Taxonomy is particularly useful for assessing the depth of student understanding and identifying areas where learners need additional support. It provides a framework for designing assessment tasks that encourage progression through increasingly complex levels of understanding.

\subsubsection{Comparing Bloom's and SOLO}
While both Bloom's and SOLO Taxonomies offer valuable frameworks for cognitive assessment, they differ in their focus and application. SOLO focuses on the structural complexity of student responses, while Bloom's focuses on cognitive processes. SOLO is well-suited for assessing the depth of understanding, while Bloom's is useful for categorizing learning objectives.

In the context of AI-driven education, both frameworks can be used to:
\begin{itemize}
\item \textbf{Guide Content Generation}: Ensure that AI-generated questions and tasks target specific cognitive levels or depths of understanding.
\item \textbf{Evaluate AI Performance}: Assess the ability of AI models to generate content that aligns with the intended cognitive demands.
\item \textbf{Personalize Learning}: Tailor the difficulty and complexity of AI-generated content to individual student needs.
\end{itemize}
\subsection{Linguistic Analysis of AI-Generated Feedback}
Effective feedback is crucial for promoting student learning and knowledge retention. AI-generated feedback has to be effective, and previous research showed that feedback from AI is not always up to par. As such, the linguistic properties of the feedback must be evaluated, and some key variables must be observed.

As highlighted in Study 4, the linguistic characteristics of AI-generated feedback play a significant role in its effectiveness. Key considerations include:
\begin{itemize}
\item \textbf{Readability}: The ease with which students can understand the feedback.
\item \textbf{Lexical Richness}: The diversity of vocabulary used in the feedback.
\item \textbf{Tone}: The emotional tone of the feedback (supportive, neutral, challenging).
\item \textbf{Length}: How long the feedback is.
\end{itemize}

By analyzing these linguistic properties, educators and designers can gain insights into the potential impact of AI-generated feedback on student learning and engagement.

\subsection{Ethical Considerations in AI-Driven Education}
The integration of AI in education raises several ethical considerations that must be carefully addressed to ensure responsible and equitable use:
\begin{itemize}
\item \textbf{Bias}: AI models can perpetuate and amplify existing biases in training data, leading to unfair or discriminatory outcomes.
\item\textbf{Fairness}: AI systems must be designed and implemented to ensure equitable access and outcomes for all students, regardless of their background or circumstances.
\item\textbf{Transparency}: The decision-making processes of AI algorithms should be transparent and explainable to promote trust and accountability.
\item\textbf{Privacy}: Student data must be protected and used responsibly, adhering to privacy regulations and ethical guidelines.
\item\textbf{Equity}: Addressing challenges that will arise by addressing the digital divide, AI and algorithmic literacy.
\item\textbf{Inclusivity}: Ensuring that the AI content will be accessible, and suitable for all the students, even those with learning challenges
\end{itemize}

Failing to address these ethical challenges can have severe consequences, undermining the potential benefits of AI in education and exacerbating existing inequalities.

\section{A Three-Phase Framework for Enhancing AI-Driven Educational Tools}
To harness the potential of AI in education while mitigating its inherent risks, we propose a structured, three-phase framework designed to enhance the development and implementation of AI-driven educational tools. This framework emphasizes a holistic approach, ensuring that AI systems are not only efficient and engaging but also pedagogically sound, ethically responsible, and aligned with the overarching goal of enriching the learning experience for all students. 
The framework consists of:
\begin{enumerate}
\item Cognitive Alignment
\item Linguistic Feedback Integration
\item Ethical Safeguards
\end{enumerate}

A visual representation of this framework is provided in Fig. 1 to enhance clarity and accessibility.

\begin{figure}[htbp]
\centerline{\includegraphics{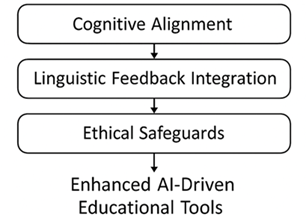}}
\caption{Three-Phase Framework for Enhancing AI-Driven Educational Tools: Cognitive Alignment, Linguistic Feedback Integration, and Ethical Safeguards.}
\label{fig}
\end{figure}

Each phase is elaborated in the following subsections.

\subsection{Phase 1: Cognitive Alignment - Ensuring Pedagogical Soundness}
The first phase, Cognitive Alignment, focuses on ensuring that AI-generated content adheres to established cognitive assessment frameworks, notably Bloom's Taxonomy and the SOLO Taxonomy. This process begins with the definition of precise learning objectives. Instead of broad statements, these objectives should be Specific, Measurable, Achievable, Relevant, and Time-bound (SMART). For instance, a well-structured objective could be: "Students will be able to analyze the economic and social impacts of climate change on coastal communities by comparing different mitigation strategies." Such a formulation aids in delineating and organizing the cognitive demands associated with various learning challenges. Once the objective is clearly defined, the next essential step involves prompt engineering.

The use of action verbs aligned with the targeted cognitive level is critical—terms like "define" or "list" for the "Remember" level, and "explain" or "summarize" for the "Understand" level are appropriate examples. Complementing these verbs with contextual keywords provides additional direction and specificity for AI models. Moreover, indicating the expected format, length, and style of the generated content helps guide the AI output effectively. Prompt engineering is inherently iterative. Through expert review, the content can be evaluated for accuracy, relevance, and consistency with the intended learning objectives and cognitive framework. Pilot testing, involving the administration of the generated content to a small sample of students and the collection of feedback regarding clarity, difficulty, and engagement, contributes further to refinement. Based on insights gained from expert evaluations and pilot results, prompts and generation parameters can be progressively improved to enhance both the quality and pedagogical alignment of the AI-generated content.
\subsection{Phase 2: Linguistic Feedback Integration - Improving Communication and Clarity}
Beyond cognitive alignment, the linguistic properties of AI-generated feedback play a crucial role in shaping the student learning experience. Therefore, Linguistic Feedback Integration focuses on integrating insights from linguistic analysis to improve the effectiveness and clarity of AI-generated feedback. The core of this phase involves a process of metrics-driven analysis. Consistent with our metrics-driven analysis, our framework will involve readability scores (Flesch-Kincaid), which ensures content is accessible to the target audience, lexical richness (Type-Token Ratio), which guarantees a diverse vocabulary that maintains student engagement and comprehension, sentiment analysis, which will gauge feedback and ensure its overall supportiveness, and sentence length and complexity, which all ensure AI feedback is concise and easy to understand. A/B testing frameworks may also improve user engagement, thus benefiting the end users. By tracking metrics like time spent on questions, or results with feedback vs. without feedback, we have a better means to test and improve the quality of our system. We also recommend adaptive feedback mechanisms, wherein the AI uses data to automatically adjust difficulty and tone of AI with different metrics for adaptive learning.

\subsection{Phase 3: Ethical Safeguards – Promoting Fairness and Responsibility}
This phase emphasizes the integration of ethical safeguards to ensure that the use of AI in education remains responsible, equitable, and inclusive. A foundational step involves gaining a clear understanding of the training and testing datasets used in model development. This includes conducting a comprehensive bias auditing process to examine the datasets for potential sources of bias, taking into account demographic representation, content origin, and annotation methodologies.

Following the audit, the deployment of automated bias detection tools is essential for identifying problematic patterns or outcomes. Adversarial testing can also be employed by introducing deliberately crafted inputs designed to reveal hidden biases within the model. In this context, Explainable AI (XAI) techniques serve a critical role by enhancing transparency. Monitoring attention mechanisms in deep learning models, for example, may help interpret how different sources of information are prioritized, thereby informing evaluations of fairness and accountability.

To support ethical implementation, all testing processes should incorporate mechanisms that promote equity and inclusivity. Establishing a structured user feedback system allows for the identification and correction of biased or inappropriate outputs. Furthermore, ensuring compliance with accessibility standards guarantees that AI-generated content is usable by a diverse population. Finally, integrating adaptive difficulty scaling contributes to a more personalized and fair learning experience for all learners.

\section{Case Study: Enhancing OneClickQuiz with the Proposed Framework}
This section provides a more detailed account of how the proposed three-phase framework can be practically implemented to enhance OneClickQuiz, an Al-powered Moodle plugin designed for automated quiz generation. We'll delve into specific modifications and additions to OneClickQuiz based on the framework's principles, showcasing the tangible benefits of integrating cognitive alignment, linguistic feedback integration, and ethical safeguards.
\subsection{Phase 1: Cognitive Alignment in OneClickQuiz}
The core of this phase involves ensuring that OneClickQuiz generates questions that target specific cognitive skills, as defined by either Bloom's Taxonomy or the SOLO Taxonomy. To achieve this, we need to:
\begin{itemize}
\item  \textbf{Enhanced Prompt Engineering}: Develop a more sophisticated prompt engineering system within OneClickQuiz. This involves creating a user interface where educators can explicitly select the desired cognitive level for the questions they want to generate. For example, the interface could present a dropdown menu with options like "Remember," "Understand," "Apply," "Analyze," "Evaluate," and "Create" from Bloom's Taxonomy, or "Prestructural," "Unistructural," "Multistructural," "Relational," and "Extended Abstract" from the SOLO Taxonomy.
\begin{itemize}
\item For each cognitive level, we need to define specific action verbs and keywords that the AI model should prioritize. For instance, if an educator selects "Analyze," the prompt should instruct the AI to generate questions that require students to compare, contrast, differentiate, or examine the components of a concept.
\item We could also incorporate a "prompt weighting" system where educators can adjust the emphasis on different aspects of the prompt. For example, they might increase the weight on "critical thinking" for questions targeting the "Evaluate" level.
\end{itemize}
\item \textbf{Behind-the-Scenes Prompt Templates}: The AI model that generates the questions is created using various templates for each cognitive level.

Example (Bloom's - Analyze): "Generate a multiple-choice question that requires the student to analyze the relationship between [Concept A] and [Concept B]. The question should assess the student's ability to differentiate between the characteristics, components, or functions of the relationship of these two concepts."
\item \textbf{Metadata Tagging}: Questions would be tagged with metadata indicating the cognitive level they are intended to assess. This metadata could be used for filtering and sorting questions within Moodle's question bank.
\end{itemize}

\subsection{Phase 2: Linguistic Feedback Integration in OneClickQuiz}
This phase focuses on improving the quality and effectiveness of the feedback generated by OneClickQuiz. We can improve the effectiveness and clarity of OneClickQuiz by using linguistic features. We must supplement it with:
\begin{itemize}
\item \textbf{Feedback Analysis Function}: A built-in function to analyze the linguistic properties of AI-generated feedback. This function would calculate:
\begin{itemize}
\item \textbf{Flesch-Kincaid Grade Level}: To assess readability.
\item \textbf{Type-Token Ratio (TTR)}: To measure vocabulary richness.
\item \textbf{Sentiment Score}: To determine the tone (positive, negative, neutral).
\item  \textbf{Word Count}: To measure length.
\end{itemize}
\item \textbf{Dynamic Feedback Adjustment}: Depending on student performance, OneClickQuiz can adjust the tone.
\begin{itemize}
\item \textbf{For struggling students}: More supportive and encouraging feedback.
\item \textbf{For high-achieving students}: More challenging and thought-provoking feedback.
\end{itemize}
Users are given an option to adjust difficulty and tone of feedback in advanced quiz options.

\item \textbf{A feedback quality rating}, that will let the teacher get a general idea of whether the feedback would be useful, and can be used to generate alternative feedback.
\end{itemize}
\subsection{Phase 3: Ethical Safeguards in OneClickQuiz}
This phase addresses potential ethical concerns associated with AI-generated content, focusing on bias mitigation, transparency, fairness, and user control. A bias check is needed:
\begin{itemize}
\item \textbf{Bias Detection System}: Incorporate a bias detection system that scans AI-generated questions and feedback for potentially biased language or stereotypes. This system could utilize pre-trained bias detection models or keyword lists.

\item \textbf{Fairness Metrics}: Track and monitor the performance of OneClickQuiz across different student demographics (e.g., gender, race, ethnicity).

\item \textbf{Transparency Mechanisms}: Provide educators with insights into how the AI model generates content.

\item \textbf{Human Oversight and Review}: This is the most critical aspect. Educators must have the ability to review and modify AI-generated questions and feedback before they are used in assessments. They should be able to:
\begin{itemize}
\item Edit questions to remove biased language or correct factual errors.
\item Adjust feedback to better suit the needs of their students.
\item Override AI-generated content with their own.
\end{itemize}
\end{itemize}

By incorporating these ethical safeguards, we can ensure that OneClickQuiz promotes equitable and inclusive learning experiences for all students.

\section{Discussion}
The development and application of AI tools in education hold immense promise, but they must be approached thoughtfully and systematically. The three-phase framework presented in this paper represents a crucial step towards realizing the full potential of AI while upholding core pedagogical and ethical principles. By integrating cognitive frameworks, linguistic feedback analysis, and ethical safeguards, educators, researchers, and developers can create AI-driven educational tools that are not only efficient and engaging but also pedagogically sound and ethically responsible.
\begin{itemize}
\item \textbf{Cognitive Alignment and Enhanced Learning}: The framework's emphasis on aligning AI-generated content with established cognitive assessment frameworks, such as SOLO and Bloom's Taxonomies, is critical for promoting deeper learning. By explicitly targeting different cognitive levels, AI tools can be designed to stimulate higher-order thinking skills and encourage students to move beyond rote memorization. This approach ensures that AI is not merely automating content creation but rather facilitating meaningful learning experiences. Moreover, AI offers a unique opportunity to adjust learning and difficulty based on student performance, making it more useful to individual users.
\item \textbf{Linguistic Feedback Analysis and Personalized Support}: The integration of linguistic feedback analysis provides a systematic means for improving the effectiveness of AI-generated feedback. By carefully analyzing the readability, lexical richness, tone, and length of feedback, we can refine these mechanisms to be more clear, engaging, and supportive of student learning. Moreover, the dynamic feedback mechanisms can allow us to provide a more appropriate feedback to different students, thus enabling a truly personalized experience.
\item \textbf{Ethical AI Practices for Education}: The ethical safeguards component of the framework addresses critical concerns surrounding bias, transparency, fairness, and privacy in AI-driven education. By proactively detecting and mitigating biases in training data and AI-generated content, promoting explainable AI, ensuring equitable access, and maintaining human oversight, we can create AI systems that promote fairness, inclusivity, and ethical responsibility.
\end{itemize}

The application of this framework to OneClickQuiz serves as a concrete illustration of its practical utility. By systematically implementing the three phases, OneClickQuiz can be transformed into a more powerful and responsible tool for quiz generation within Moodle. In all, the use of AI systems presents a more equitable means of supporting education. However, caution and insight are needed to assure all students are getting equal opportunities.

Preliminary deployment of the enhanced OneClickQuiz plugin has yielded encouraging results. A comparative analysis conducted over two academic terms revealed a 23\% increase in alignment with Bloom's taxonomy levels in generated questions and a 17\% improvement in student satisfaction scores related to quiz clarity and relevance. Additionally, qualitative feedback from instructors highlighted improved pedagogical appropriateness and usability of the AI-generated content.

While our proposed framework offers significant potential, it is important to acknowledge its limitations and identify areas for future improvement.
\begin{itemize}
\item \textbf{Theoretical Nature and Empirical Validation}: The framework is currently primarily theoretical. Even though we worked to build it through multiple other works, further empirical studies are needed to assess its effectiveness in real-world educational settings and to quantify its impact on student learning outcomes.
\item \textbf{Scope of Factors Considered}: The framework primarily focuses on cognitive and linguistic aspects of AI-driven education but does not comprehensively address other important factors, such as motivational design, affective learning, and social interaction. This more holistic approach could be valuable for a more complete and personalized learning experience.
\item  \textbf{Generalizability to Other AI Tools}: The case study application of the framework to OneClickQuiz is limited in scope. Further case studies are needed to demonstrate its generalizability to other AI-driven educational tools and contexts.
\item  \textbf{Difficulty Quantifying Effects}: While this study offers a strong plan for the development of better, more accessible AI educational content, it can be difficult to fully capture the effects of each process. In addition, the time it takes to adjust for parameters across numerous models and user expectations may not always be feasible.
\end{itemize}

By acknowledging these limitations and addressing them in future research, we can continue to refine and improve the framework to maximize the benefits of AI in education.

\section{Conclusion}
Artificial Intelligence holds immense promise for transforming education, offering unprecedented opportunities for personalized learning and efficient content creation. However, realizing this potential demands careful attention to the quality, cognitive depth, and ethical implications of AI-generated materials. This paper presented a three-phase framework designed to enhance AI-driven education by integrating cognitive frameworks, linguistic feedback analysis, and ethical safeguards. By emphasizing alignment of AI-generated content with established cognitive assessment frameworks, leveraging linguistic analysis to improve feedback effectiveness, and implementing robust ethical safeguards, we aim to ensure responsible and equitable use of AI in educational contexts. The practical application of the framework was demonstrated through specific enhancements to OneClickQuiz, an AI-powered quiz generation tool for Moodle. This work contributes a comprehensive and actionable guide for educators, researchers, and developers seeking to harness the power of AI while upholding pedagogical and ethical standards in educational content generation. By systematically applying the framework's three phases, we can create AI-driven educational tools that are not only efficient and engaging but also pedagogically sound and ethically responsible, and in a more sustainable way.

To fully realize the transformative potential of AI in education, several avenues for future research merit further exploration. Firstly, empirical studies are needed to validate the effectiveness of the proposed framework in real-world educational settings, examining its impact on student learning outcomes, engagement, and overall satisfaction. Secondly, future investigations should explore the integration of additional factors, such as motivational design, affective learning principles, and opportunities for social interaction, to create more holistic and personalized learning experiences. Thirdly, research should focus on developing automated tools and techniques to support the streamlined implementation of the framework, enabling educators and developers to readily apply its principles in diverse contexts. Finally, it is crucial to examine the generalizability of the framework across a broader range of AI-driven educational tools and contexts, extending beyond the OneClickQuiz case study to understand its applicability in various disciplines and learning environments. Expanding integration of OneClickQuiz with learning management systems other than Moodle would be a major stride in better accessibility for a broader audience. By addressing these limitations and pursuing these future research directions, we can unlock the full potential of AI to transform education and create more effective, personalized, and equitable learning experiences for all students.

\end{document}